\documentclass[10pt, a4paper]{article}

\usepackage[]{lrec-coling2024} 
\usepackage{multirow}
\usepackage{tabularx}
\usepackage{booktabs}
\usepackage{times}
\usepackage{latexsym}
\usepackage{adjustbox}
\usepackage{graphicx}
\usepackage{color}
\usepackage{colortbl}
\usepackage{adjustbox}
\usepackage{soul}

\title{BanglaQuAD: A Bengali Open-domain Question Answering Dataset}

\name{Md Rashad Al Hasan Rony$^1$, Sudipto Kumar Shaha$^1$, Rakib Al Hasan$^1$\\ {\bf \large Sumon Kanti Dey$^1$, Amzad Hossain Rafi$^1$, Amzad Hossain Rafi$^1$} \\ {\bf \large Ashraf Hasan Sirajee$^1$, Jens Lehmann\thanks{$^*$Work done outside of Amazon.}$^{2,3, *}$}} 

\address{BanglaAI$^1$, TU Dresden$^2$, Amazon$^3$}

\abstract{
Bengali is the seventh most spoken language on earth, yet considered a low-resource language in the field of natural language processing (NLP). Question answering over unstructured text is a challenging NLP task as it requires understanding both question and passage. Very few researchers attempted to perform question answering over Bengali (natively pronounced as \textit{Bangla}) text. Typically, existing approaches construct the dataset by directly translating them from English to Bengali, which produces noisy and improper sentence structures. Furthermore, they lack topics and terminologies related to the Bengali language and people. This paper introduces BanglaQuAD, a Bengali question answering dataset, containing 30,808 question-answer pairs constructed from Bengali Wikipedia articles by native speakers. Additionally, we propose an annotation tool that facilitates question-answering dataset construction on a local machine. A qualitative analysis demonstrates the quality of our proposed dataset.
 \\ \newline }

\begin{document}

\maketitleabstract

\section{Introduction}

Question answering (QA) over unstructured text is challenging, as the system requires understanding of a wide range of vocabulary and question types. Machine reading comprehension (MRC) is an extractive question answering task that focuses on detecting a continuous answer span within a passage for a given question. 
Dataset plays a key role in developing such QA systems. A dataset that covers a wide range of vocabulary, question types, and variable answer lengths can facilitate the system with more textual patterns to learn. 


Although question answering over English text is a widely studied topic~\cite{rajpurkar-etal-2016-squad,rajpurkar-etal-2018-know,ijcai2022p729}, only a few research works focus on Bengali question answering~\cite{tahsin2021deep,bhattacharjee-etal-2022-banglabert}. The lack of a high-quality Bengali QA dataset is one of the primary reasons for this. Existing Bengali QA systems use neural translators~\cite{devlin-etal-2019-bert} to transform an English QA dataset into Bengali to obtain training data~\cite{bhattacharjee-etal-2022-banglabert,aurpa2022reading}. These types of data sets suffer from three major issues. \textbf{Firstly}, the meaning of the translated Bengali text is often incorrect due to the use of a neural translator. \textbf{Secondly}, they often contain grammatically and structurally incorrect text. \textbf{Thirdly}, they lack terminologies from the Bengali domain (i.e., topic, place, and individuals), because English datasets are built from articles largely related to the English speaking world.
Addressing the shortcomings, we present an open-domain Bengali question answering dataset, BanglaQuAD. BanglaQuAD contains 30,808 high-quality human-annotated question-answer pairs. Specifically, we select 658 articles from more than 12,000 Bengali Wikipedia articles based on importance and frequently encountered topics. We follow the official Wikipedia categories of articles and dive deep one level into sub-categories to further widen the scope of the dataset. This selection process ensures a broader coverage of topics, resulting in a wide range of vocabulary from different topics. Human annotators are employed to construct the question-answer pairs from passages of the curated articles. This ensures that the data is of the best possible quality. To diversify and extend the understanding of QA systems, BanglaQuAD includes diverse question types, unanswerable questions, answers with variable lengths. Furthermore, we introduce a Bengali annotation tool, \textit{BnAnno}, by extending the cdQA~\footnote{\url{https://github.com/cdqa-suite/cdQA-annotator}} tool to construct a Bengali question answering dataset. \textit{BnAnno} supports unstructured text and converts them into the popular SQuAD dataset format after the annotation. The SQuAD dataset format is widely adopted for developing QA systems~\cite{bhattacharjee-etal-2022-banglabert}. The user interface of \textit{BnAnno} is developed in Bengali to assist native Bengali annotators. The contributions of this paper are summarized as follows:
\begin{itemize}
        \item A high quality human-annotated Bengali question-answer dataset, \textit{BanglaQuAD}, suitable for machine reading comprehension and information retrieval tasks. BanglaQuAD contains 30,808 question-answer pairs, annotated by native Bengali speakers.
        \item An annotation tool, \textit{BnAnno}, to construct question answer pairs from unstructured text data, which would encourage further research on Bengali question answering systems.
    \end{itemize}

BanglaQuAD can be used as a benchmark for developing question answering and information retrieval systems on Bengali text. We made the dataset and annotation tool publicly available~\footnote{\url{https://github.com/rashad101/BanglaQuAD-LREC-COLING-24}}.

\section{BanglaQuAD}
\subsection{Dataset Generation Workflow}
\begin{figure}[!ht]
\centering
    \includegraphics[width=1.02\columnwidth]{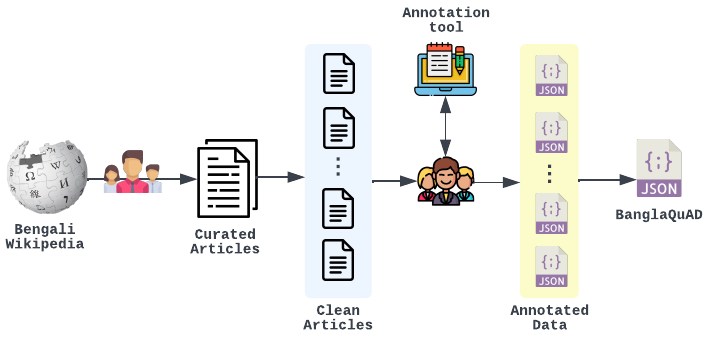}
    \caption{Dataset generation workflow.} 
    \label{fig:pipeline}
\end{figure}

To construct BanglaQuAD, first, we assign three human annotators to shortlist 658 Wikipedia articles depending on their importance and the likeliness of a question being asked by a Bengali speaking person. To eliminate noisy text, the curated articles are passed through a cleaning process before starting the annotation process. A group of annotators construct question answer pairs based on the assigned articles using \textit{BnAnno} (\textsection{\ref{sec:anno}}). The data set construction pipeline is depicted in Figure~\ref{fig:pipeline}. The following subsections provide a brief description of the dataset construction process.

\begin{figure}[!ht]
\centering
        \includegraphics[width=1.03\columnwidth]{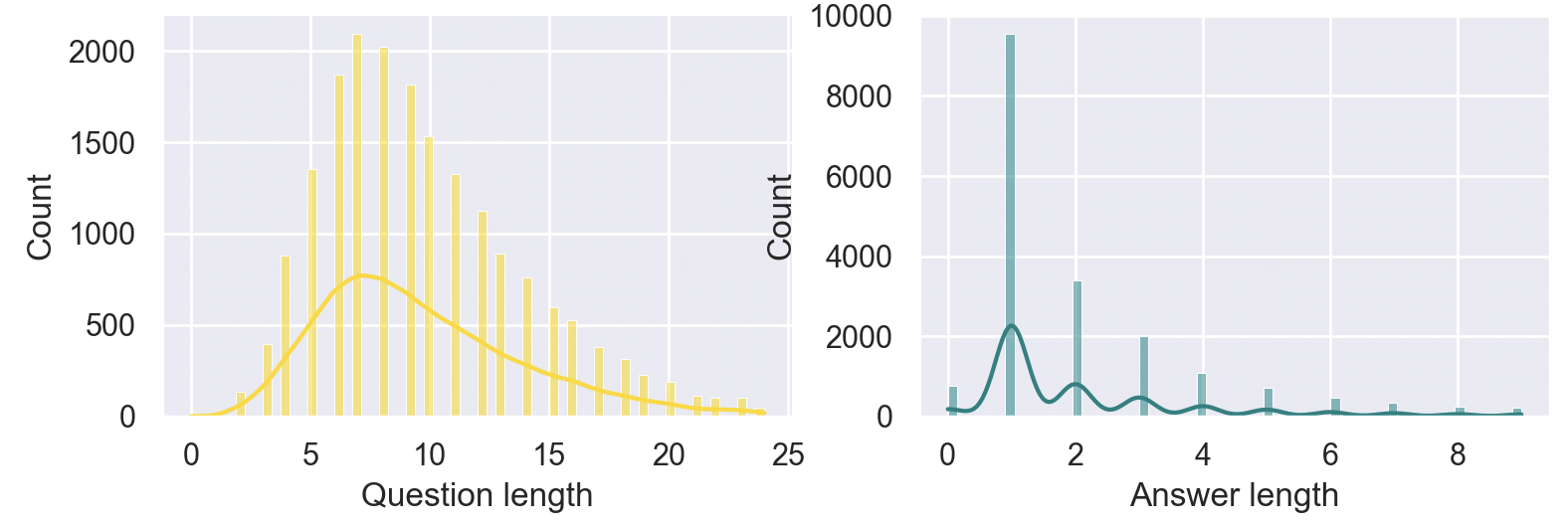}
    \caption{Question and answer lengths distribution.} 
    \label{fig:qadist}
\end{figure}
\paragraph{Data Collection:}
BanglaQuAD is developed based on Bengali Wikipedia articles\footnote{\url{https://bn.wikipedia.org/wiki/}}. We divided Bengali Wikipedia articles into 14 categories and 114 subcategories following the official topic types. Annotators directly copied the content of the article into the annotation tool to start the annotation process. To eliminate noisy text references (i.e., [3] and [3,5]) and an English texts are removed from the text. 

\paragraph{Data Annotation:}
We instruct seven native Bengali-speaking human annotators (four from Computer Science (CS) and three from non-CS background) to first construct questions from a set of passages and then select an answer span in the passage that answers the question. We uniformly distributed 658 articles among the annotators. The articles are selected based on their importance and frequency. Next, the annotators utilized our annotation tool to construct question-answer pairs by following the annotation guidelines listed below:
\begin{itemize}
    \item Please go through each sentence in a given passage and create only factual question-answer pairs whenever possible. If there exist multiple facts in a given sentence, then please create multiple question-answer pairs for each fact in that.
    \item Do not ask questions that require a summary of the text to answer. Only ask questions where the answer is a continuous span in the given passage.
    \item At least construct five unanswerable questions from each article.
\end{itemize}
\begin{figure*}[!ht]
\vspace{-0.5cm}
\centering
\includegraphics[width=1\textwidth]{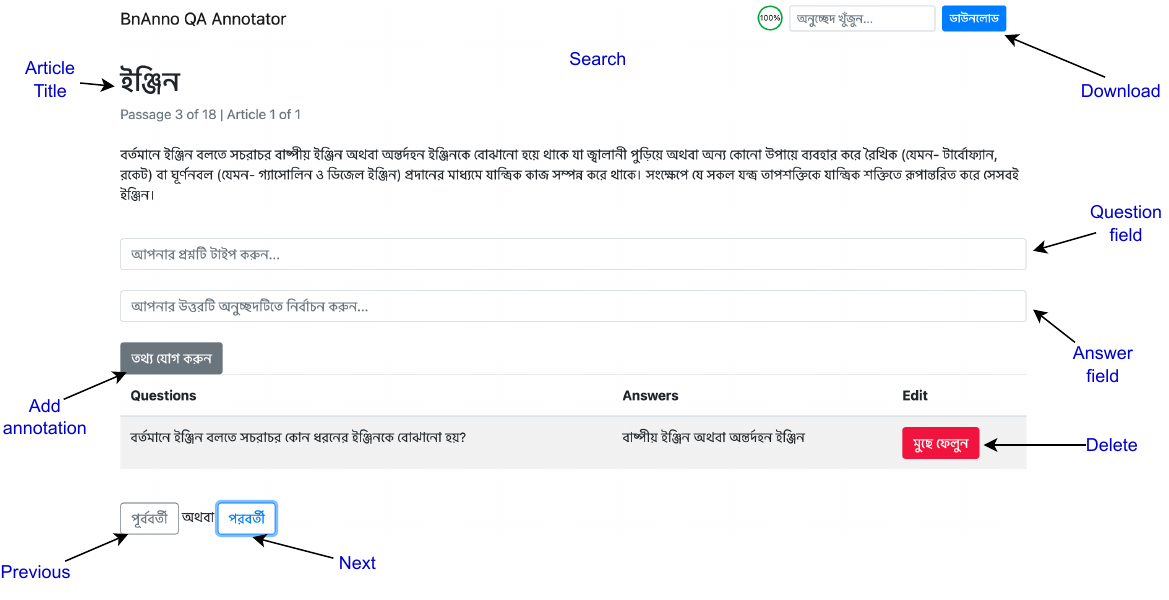}
    \caption{A snippet of the annotation tool. English translations are provided in blue text for demonstration.} 
    \label{fig:anno}
\vspace{-0.2cm}
\end{figure*}
A JavaScript Object Notation (JSON) file containing the annotated data in SQuAD format is generated by the annotation tool for each article upon completion. Finally, we combine all the generated JSON files into a single file to obtain the final version of BanglaQuAD.

\begin{table}[]
\centering
\begin{adjustbox}{width=0.8\columnwidth,center}
\begin{tabular}{lr}
\toprule
 & \textbf{Count} \\\midrule
Total number of categories     & 14     \\
Total number of sub-categories & 114     \\
Total number of articles       & 658    \\
Total number of passages       & 6,425       \\
Number of QA-pairs             & 30,808 \\
Unanswerable questions         &  1,071    \\
Average passage length &  95.73      \\
Average question length &   9.43     \\
Average answer length  &   3.08    \\
Vocabulary Size          &     87,482  \\\bottomrule
\end{tabular}
\end{adjustbox}
\caption{Dataset statistics.}
\label{tab:datastat}
\vspace{-0.4cm}
\end{table}
\subsection{BnAnno: A Bengali Annotation Tool}
\label{sec:anno}

In this work, we extended the cdQA annotation tool to support Bengali language. The standard cdQA tool requires the users to pre-process the data into SQuAD-like format in order to use the tool. In contrast, in BnAnno the user can copy any text document to the tool and start the annotation right away without having to think about pre-processing and the output format. This saves a lot of time and effort in obtaining the dataset, which is essential for constructing large-scale datasets. Figure~\ref{fig:anno} depicts a snippet of the annotation tool. 

\begin{figure}[]
\centering
        \includegraphics[width=1.04\columnwidth]{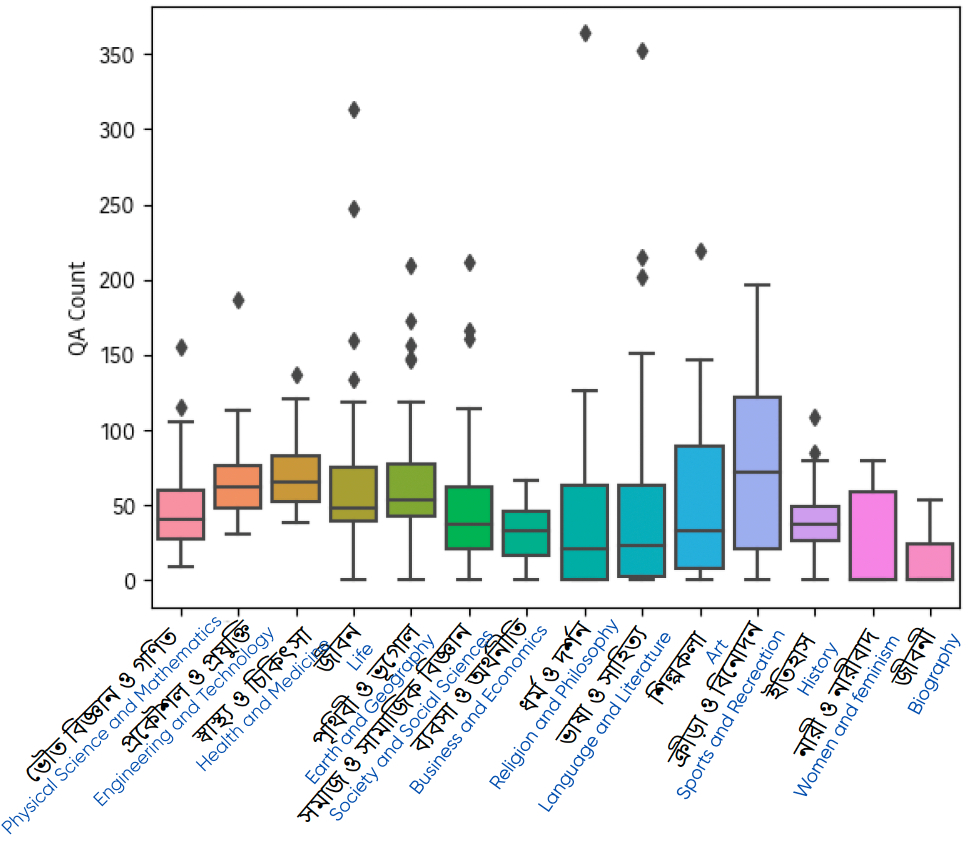}
    \caption{QA pairs per category. For demonstration purpose English translations are in blue text.} 
    \label{fig:catcount}
    \vspace{-0.4cm}
\end{figure}
\subsection{Dataset Statistics: } BanglaQuAD covers a wide range of general topics besides the Bengali domain, depicted in Figure~\ref{fig:catcount}. Furthermore, Table~\ref{tab:datastat} shows core dataset statistics. On average, about 47 questions were asked from each article, indicating the depth and coverage of various types of facts and related questions. Figure~\ref{fig:qadist} depicts the distribution of question and answer lengths. A wide range of question and answer lengths makes the dataset very challenging. Our intuition is that it would stress the capabilities of QA systems. Furthermore, BanglaQuAD has a vocabulary size of 87,482, spanning 14 major topics that include life and Bengali language related domains: Physical Science and Mathematics, Engineering and Technology, Health and Medicine, Life,  Earth and Geography,  Society and Social Sciences, Business and Economics, Religion and Philosophy, Language and Literature, Art, Sports and Recreation, History, Women and feminism, Biography, and India. This would enable the QA system to handle a wide range of contexts. We split the dataset into 80-20 train-test split to obtain 24,646 train and 6,162 test data.

\section{Experiments and Analysis}

\paragraph{Experimental Setup:} We select baseline models that are specifically trained on Bengali language: BanglaBERT~\cite{bhattacharjee-etal-2022-banglabert} and  IndicBERT~\cite{kakwani-etal-2020-indicnlpsuite}. BanglaBERT is a pre-trained BERT~\cite{devlin-etal-2019-bert} model trained on Bengali text where IndicBERT is a multilingual model trained on Indian language. The baseline models are trained with the recommended hyper-parameters.

As benchmark dataset we select publicly available Bengali QA datasets:  UDDIPOK~\cite{aurpa2023uddipok} and TYDI QA~\cite{clark-etal-2020-tydi}. UDDIPOK is a crowd sourced question answering dataset collected from newspaper, textbook and exam questions. The dataset contains 2,908 training and 728 test samples. TYDI QA is a non-translation based crowd sourced multilingual question answering dataset. It contains 10,768 training and 334 test data. We develop BanglaQuAD a machine reading comprehension dataset constructed based on Bengali Wikipedia article. The dataset contains 24,646 training and 6,162 test data.
\begin{table}[]
\begin{adjustbox}{width=0.9\columnwidth, center}

\begin{tabular}{lllll}
\toprule
\multirow{5}{*}{} & \multicolumn{2}{c}{\textbf{BanglaBERT}}         & \multicolumn{2}{c}{\textbf{IndicBERT}}          \\ \cline{2-5} 
                  & \multicolumn{1}{c}{\textbf{EM}} & \multicolumn{1}{c}{\textbf{F1}} & \multicolumn{1}{c}{\textbf{EM}} & \multicolumn{1}{c}{\textbf{F1}} \\\midrule
UDDIPOK           &        63.44                &      70.35                  &          59.92              &        62.60                \\
TYDI QA            &     32.42                   &       42.84                 &          30.10              &          47.58              \\
BanglaQuAD        &        22.92                &           39.38           &            21.50            &          37.22             \\\bottomrule
\end{tabular}
\end{adjustbox}
\caption{Performance of baseline models.}
\label{tab:exp}
\vspace{-0.4cm}
\end{table}
\begin{figure*}[!ht]
\vspace{-0.5cm}
\centering
        \includegraphics[width=0.78\textwidth]{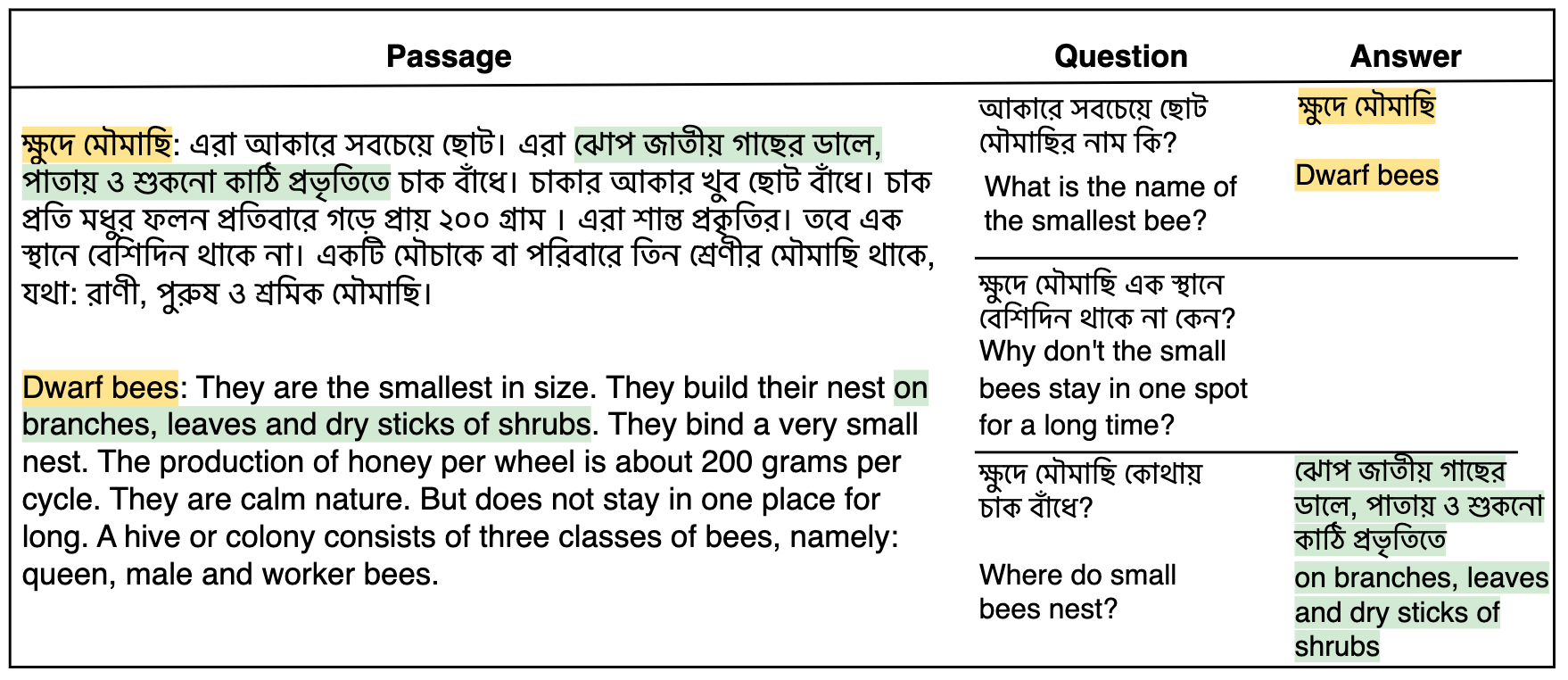}
    \caption{Example annotated QA pairs. English text is provided for explanation purpose. Answer spans are highlighted in yellow and green.} 
    \label{fig:exmp}
\vspace{-0.45cm}
\end{figure*}
\paragraph{Quantitative Analysis:} 
Table~\ref{tab:exp} reports the results in Exact Match (EM) and F1 score. In both cases on the BanglaQuAD dataset the baseline models obtained the lowest score. This result exhibits that pre-trained models fine-tuned on BanglaQuAD find it difficult to understand diverse range of topics and question types. This characteristics of BanglaQuAD would stress the comprehension capabilities of QA models and help develop robust QA systems.

\paragraph{Qualitative Analysis:}
Figure~\ref{fig:exmp} demonstrates three example question-answer pairs based on a given passage. The examples of QA pairs include three types of \textit{Wh}-questions (\textit{what}, \textit{why}, and \textit{where}), leading to answers with variable lengths. To challenge the development of QA systems, we include questions where the answers do not exist in the given paragraph. The second example question exhibits such a scenario. BanglaQuAD contains nine types of \textit{Wh}-questions. The distribution is as follows: \textit{what} (13\%), \textit{when} (12 \%), \textit{where} (12 \%), \textit{who} (7 \%), \textit{whom} (6 \%), \textit{which} (6 \%), \textit{whose} (5 \%), \textit{why} (6 \%), and \textit{how} (4 \%).  The rest of the 29\% questions start with a keywords or names. A QA system trained on a wide variety of question types would improve the systems' language understating capabilities and question-answering performance.

To assess the quality of the annotation process, we randomly chose 100 data points and asked two annotators to select an answer span, given a question and a passage. The inter-annotator agreement (Cohen's kappa $\kappa$) of the answer span selection experiment is 0.79).

\section{Related Work}

Machine reading comprehension is a widely adopted form of question answering over unstructured text~\cite{chen-etal-2017-reading,guu2020retrieval,yang-etal-2019-end-end,raffel2020exploring}. Recent large-scale MRC datasets contributed to the rapid development of MRC systems~\cite{xiong-etal-2019-tweetqa,jing-etal-2019-bipar,xiong-etal-2019-tweetqa,rajpurkar-etal-2016-squad,ijcai2022p729}. The most popular one is SQuAD~\cite{rajpurkar-etal-2018-know} introduces 100,000 open-domain English question-answer pairs. In a different work, an investigation~\cite{dzendzik2021english} on 60 English MRC datasets reveals that Wikipedia is commonly used as a source of unstructured data. Furthermore, they suffer from a deficiency of question types beginning with \textit{why}, \textit{when}, and \textit{where}.
    
    

Bengali MRC systems typically translate English data into Bengali for developing question answering systems~\cite{bhattacharjee-etal-2022-banglabert,aurpa2022reading,9528178}. Very recently,~\cite{bhattacharjee-etal-2022-banglabert} trained a BERT~\cite{devlin-etal-2019-bert} model on a translated SQuAD dataset to develop an open domain QA system. In a different work,~\cite{aurpa2022reading} translated 3,675 question-answer pairs from a COVID-19-related English MRC dataset into Bengali. A non-translation based Multilingual TYDI QA~\cite{clark-etal-2020-tydi} is also proposed where only 11,430 Bengali QA-pairs are available. 
In contrast to the existing datasets, we focus on maintaining quality in addition to quantity. We select articles based on importance and based on the Bengali domain (i.e., topic, place, and individuals). We instruct human annotators to ask diverse type of questions, leading to factual and variable-length answers.

\section{Conclusion}
We have presented BanglaQuAD, a Bengali question answering dataset containing 30,808 high-quality human-annotated question-answer pairs. Furthermore, we developed an annotation tool to construct question-answer pairs from unstructured text data. We made the dataset and annotation tool publicly available to promote research on the Bengali language and text. In the future, we plan extend the dataset and create a question answering public leader-board based on BanglaQuAD to encourage further research on open-domain question answering. 

\section{Ethical Considerations}
We carefully checked and confirm that the work conducted in this paper does not violate any ethical considerations.

\section{Bibliographical References}\label{sec:reference}

\bibliographystyle{lrec-coling2024-natbib}
\bibliography{lrec-coling2024-example}

\bibliographystylelanguageresource{lrec-coling2024-natbib}

\end{document}